\documentclass[conference, a4paper]{IEEEtran}
\IEEEoverridecommandlockouts

\ifCLASSINFOpdf
  \usepackage[pdftex]{graphicx}
  \DeclareGraphicsExtensions{.pdf,.jpeg,.png}
\else
\fi

\usepackage[minnames=1, maxnames=6]{biblatex}
\usepackage{amsmath,amssymb,amsfonts}
\usepackage{algorithmic}
\usepackage{graphicx}
\usepackage{tabularx}
\usepackage{textcomp}
\usepackage[left=1.4cm, right=1.4cm, top=1.7cm]{geometry}
\usepackage{caption}
\usepackage{xcolor}
\usepackage{anyfontsize}
\usepackage{placeins}
\usepackage{balance}
\makeatletter
\newcommand*\titleheader[1]{\gdef\@titleheader{#1}}
\AtBeginDocument{%
\let\st@red@title\@title
\def\@title{%
\bgroup\normalfont\normalsize\centering\@titleheader\par\egroup
\vskip0.2em\st@red@title}
}
\makeatother

\makeatletter
\renewcommand{\fnum@figure}{Figure \thefigure}
\makeatother

\title{ {Graph-based Complexity Forecasts in UK En Route Airspace Using Relevant Aircraft Interactions} \\
\thanks{This work is supported by the grant ``EP/V056522/1: Advancing Probabilistic Machine Learning to Deliver Safer, More Efficient and Predictable Air Traffic Control'' (aka Project Bluebird), an EPSRC Prosperity Partnership between NATS, The Alan Turing Institute, and the University of Exeter.}
\vspace{0.5cm}
}

\titleheader{Second US-Europe Air Transportation Research and Development Symposium (ATRDS2026)}

\author{\IEEEauthorblockN{Edward Henderson}
\IEEEauthorblockA{The Alan Turing Institute \\
London, England \\
\textbf{ehenderson@turing.ac.uk}}
\and
\IEEEauthorblockN{George {De Ath}}
\IEEEauthorblockA{University of Exeter \\
Exeter, England \\
\textbf{g.de.ath@exeter.ac.uk}}
\and
\IEEEauthorblockN{Nick Pepper}
\IEEEauthorblockA{The Alan Turing Institute \\
London, England\\
\textbf{npepper@turing.ac.uk}}
}

\renewcommand\IEEEkeywordsname{Keywords}
\IEEEaftertitletext{\vspace{-1\baselineskip}}

\begin{document}

\maketitle

\noindent \begin{abstract}

Effectively managing Air Traffic Control Officer (ATCO) workload is crucial in maintaining operational safety. Group supervisors currently use tools that estimate upcoming traffic load to aid decision-making. However, industry-standard models can fail to capture the nuances of upcoming air traffic complexity. This study presents a probabilistic approach to forecast the complexity of an airspace sector using the number of \textit{relevant aircraft pairs}, i.e., those that require monitoring or deconfliction by a controller, as a proxy measure for ATCO workload. We adapted an existing relevant aircraft filter algorithm to make it suitable for use in London Middle Sector (LMS), a complex airspace sector with multiple climbing and descending flows of traffic above some of the busiest airports in Europe. Through iterative feedback with licensed ATCOs, the algorithm was refined and extended to handle specific geometric and operational considerations. The updated algorithm outperformed the original, with an F1-score of 0.84 compared to 0.69 on an ATCO-labelled set of 50 traffic scenarios.
To produce forecasts of future numbers of relevant aircraft pairs in the sector, a resampled graph representation of the route network through LMS was constructed, standardising the spatial fidelity of route legs across the sector.
The forecasting method accounts for uncertainty in aircraft arrival times by modelling the probability of each aircraft occupying route segments at future query times.
When combined with historic distributions of relevant interactions and a live operational data stream, predictions of upcoming ATCO workload could be made up to 45 minutes in advance. The proposed method to forecast upcoming workload showed a significantly stronger correlation with actual relevant interactions (Spearman's $\rho = 0.68$) than a standard traffic volume prediction ($\rho = 0.55$). The resulting data-driven tool shows promise for use by group supervisors to inform sector configuration and ATCO rostering decisions.

\end{abstract}

\vspace{0.3cm}

\begin{IEEEkeywords}
Air traffic management;
Airspace complexity;
Graph-based representation;
Complexity forecasting;
Time series forecasting;
Decision support tool;
Air transportation
\end{IEEEkeywords}

\section{Introduction}

A key responsibility of group supervisors is to manage the workload of Air Traffic Control Officers (ATCOs) in their unit, deciding on rostering and the configuration of the sectors for which they are responsible \cite{group_supervisor}. The workload of ATCOs is driven by the complexity of air traffic within the sector~\cite{Djokic2010, vanPaassen2010}. Therefore, accurately predicting upcoming complexity plays a crucial role in informing the decisions of the group supervisor. 

However, this task is challenging both in terms of quantifying complexity and its accurate prediction. Complexity heavily depends on the operational context of the sector~\cite{Eurocontrol_complex}. Each sector has a unique geometry, set of procedures, and traffic flows, with units of ATCOs specialising on particular sectors. Therefore, numerical measures of complexity are not necessarily generalisable across sectors~\cite{AntulovFantulin2020}. Complexity is sector-specific and can only be directly measured through the perceived workload of ATCOs. Furthermore, the task of forecasting complexity is hampered by the inherent uncertainty within the air traffic system. System-wide disruptions, such as convective weather cells or major delays at airports \cite{PYRGIOTIS201360}, can arise unexpectedly and create significant increases in complexity \cite{delaura2008modeling, CV_uncertainty}. The situational complexity within individual sectors is highly sensitive to the arrival time of aircraft in a sector. For instance, a delay of only a few minutes might cause a pair of aircraft to require deconfliction by an ATCO that might not have otherwise occurred. 

Given these confounding factors, industry-standard methods use heuristic measures of complexity such as the number of aircraft in a sector or the number of climbing and descending aircraft, with sector-specific hotspot factors that attempt to coarsely account for flows of traffic within a sector. These models can provide a lookahead time on the order of hours but are lacking in several respects. Firstly, the underlying trajectory prediction (TP) is deterministic and cannot capture deviations that will arise from the significant levels of epistemic uncertainty concerning an aircraft's performance. Industry-standard models measure complexity through the occupancy of specific routes and do not have the fidelity to model the future traffic configuration within the sector~\cite{nats2010altran, ed_complexity}. This leads to over-estimation of complexity in cases where traffic levels within sectors are low, but where several aircraft happen to traverse these routes. Similarly, the assumption of these models is that aircraft will follow their filed route throughout the sector. However, in practice, aircraft are likely to be vectored and enter subsequent sectors off route, particularly if convective weather cells are forecast. Given these limitations, in this paper we argue that a \textit{probabilistic approach} that leverages both historic distributions of traffic data and livestreamed operational data is required. Such a model has the capability to provide a more accurate forecast by learning trends in historic data, with the capacity to refine its predictions in real-time as the operational situation evolves.

Central to our approach for complexity forecasting is the concept of \textit{relevant aircraft} as a proxy measure of complexity \cite{VijayKumbhar2024,Liu2026, Mohamed2026}. We use this term expansively to cover not only pairs of aircraft that are in conflict and will require deconfliction, but also pairs of aircraft that may require close monitoring by the ATCO but without the need for deconfliction. This requires a measure of relevancy that is by definition much more conservative than separation standards.

Using the number of relevant pairs of aircraft as a proxy for complexity, the task is then to predict the number of relevant pairs of aircraft within a sector over a specified time horizon. Building on previous academic work, which has leveraged graph theory to flexibly model the geometry of an airspace or traffic sample, we develop a graph-based representation of a specific sector (see, e.g., \cite{ed_complexity, LI2024104521,marzuoli2011two, PANG2023102113, imperial_graph}). Extending this literature, we use arrival time estimates from live operational data and a distribution of relevant pairs identified from a large historic dataset of traffic data to make predictions of upcoming relevant traffic. Crucially, the presented method accounts for uncertainties in the arrival time estimates, which are inherent in the real-world system. This work makes several contributions to the literature: 
\begin{itemize}
    \item A \textbf{relevant aircraft filter} that has been tuned to a specific sector in the London Area Control Centre using \textbf{ATCO feedback};
    \item A \textbf{graph-based, probabilistic model} for forecasting numbers of relevant aircraft pairs within a sector of airspace;
    \item \textbf{Improved prediction} of relevant aircraft pairs, used as a proxy for complexity, over a 45-minute lookahead time. 
\end{itemize}

\section{Methodology}

\subsection{London Middle Sector (LMS)}

We focus on developing a relevant aircraft filter and complexity prediction method for a specific sector within the London Area Control Centre (LACC) environment. The chosen sector, London Middle Sector (LMS), has an altitude range between flight level (FL\footnote{altitude expressed in increments of 100 feet.}) 215 and 305 and sits above the London Terminal Manoeuvring Area, which includes departures and arrivals at several of Europe's busiest airports. The altitude range of the sector is such that the majority of the traffic in LMS is either climbing or descending, with climbing aircraft typically flowing southwards and eastwards towards sectors higher up in the LACC. On the other hand, the majority of traffic flying north or westwards are beginning their descents towards international airports in the Midlands, the west of England, and Wales. 

The dataset used to tune the relevant aircraft filter for LMS comprised radar tracks from transponder-based secondary radar, scheduled flight information, coordination data detailing the hand-off agreements made by ATCOs to pass aircraft between airspace sectors, and the clearances, or instructions, issued to pilots by ATCOs. Latitude and longitude positions for the navigation waypoints contained within the LMS sector are stored in the Project Bluebird digital twin of LACC, which provided an environment within which the presented model could be prototyped~\cite{Pepper2026}.

\subsection{Relevant aircraft filter adaptation for LMS}
\label{meth:adaptation}
Vijay Kumbhar~\textit{et al.} introduced a flight filtering algorithm, designed as a decision support tool for ATCOs, that identifies flights relevant to a selected subject aircraft on their radar display based on spatio-temporal interactions~\cite{VijayKumbhar2024}. The authors developed their filtering method for the Brussels West airspace sector using a set of 36 static traffic scenarios. Each scenario had a single flight of interest, or subject aircraft, highlighted, and a group of ATCOs annotated which other flights in the sector were relevant to the subject aircraft. The parameters of their filtering algorithm were tuned to align with these ATCO annotations and are shown in Table~\ref{tab:original_parameters}.

\begin{table}[ht]
\centering
\caption{\label{tab:original_parameters} Overview of the parameters obtained by Vijay Kumbhar~\textit{et al.} for the Brussels West airspace sector for their relevant aircraft filtering approach. Updated parameters for LMS and our proposed method are shown in the final column. The $t_{CPA}$ parameter was redundant in the updated approach after removal of the state-based mode. Table adapted from~\cite{VijayKumbhar2024}.}
\begin{tabularx}{\columnwidth}{c X c c}
Parameter & Description & Original & LMS tuned \\ \hline
\noalign{\vspace{1ex}}
$\Delta FL$ & Vertical distance threshold, the distance between the spans of the current (cleared), selected and exit flight levels (increments of 100~ft) of each aircraft to determine whether pairs of aircraft could be relevant. & 0 FL & 10 FL \\
\noalign{\vspace{0.5ex}}
$d$ & Current lateral distance between aircraft. & 160 NM & 80 NM \\
\noalign{\vspace{0.5ex}}
$\Delta T$ & Lookahead time for trajectory predictions. & 11 mins & 12 mins \\
\noalign{\vspace{0.5ex}}
$d_{CPA}$ & Minimum lateral distance (closest point of approach, CPA) between two aircraft within the lookahead window. & 12 NM & 15 NM \\
\noalign{\vspace{0.5ex}}
$t_{CPA}$ & Maximum time for aircraft to reach the CPA for state-based trajectory prediction. & 11 mins & N/A \\
\end{tabularx}
\end{table}

The original relevant aircraft filter method of Vijay Kumbhar~\textit{et al.} uses TP to forecast the future positions of aircraft over a specified lookahead time and then implements a series of thresholds and rules to determine which aircraft are relevant to a subject aircraft. Their filtering method consists of two main branches, one using state-based and the other intent-based trajectory predictions. The state-based filter compares future trajectories predicted under the assumption aircraft follow their present speeds and headings, without considering the intended flight plans. The intent-based filter instead compares predicted trajectories following each aircraft's intended route. The state-based element is intended to highlight more imminent, critical interactions, whereas the intent-based component captures relevant aircraft on a longer time scale.

Recognising that sectors have unique combinations of geometry, procedures, and traffic flows, which can limit the generalisability of automated tooling across sectors, two licensed ATCOs from the Swanwick Centre were consulted to gather qualitative evidence on the efficacy of the relevant aircraft filter when applied to LMS. Feedback was collected in several rounds. In the first round, ATCOs were shown the outputs of the Vijay Kumbhar~\textit{et al.} filter, using the parameter values listed in the third column of Table~\ref{tab:original_parameters}. Subsequently, the parameter values of the filter were tuned and additional logic was incorporated in response to the ATCO feedback, with the right-hand column of Table~\ref{tab:original_parameters} displaying the final parameter values for adaptation of the filter to LMS.

Specifically, during each cycle of feedback, ATCOs were presented with a survey containing 50 static traffic samples. Each sample contained a single subject aircraft and aircraft that were judged to be relevant to the subject by the filter were highlighted. Samples were selected from time periods with relatively high traffic levels (6 - 23 aircraft, with a median of 15) during August 2019, at a point where the subject aircraft had recently entered the sector. Subject aircraft were selected with a wide range of destinations and routes to promote diversity in the traffic samples. The ATCOs were instructed to treat the traffic samples as if they were controlling the entire LMS sector, even though the airspace may have been band-boxed or sub-divided and controlled by two separate ATCOs at the time. This allowed us to introduce scenarios with very high traffic density and extract as many candidate pairs as possible out of each traffic sample in the survey.

When completing the survey, the ATCOs were asked to identify any relevant aircraft missed by the filter, as well as those that were incorrectly flagged as relevant to subject aircraft. As the concept of relevant aircraft can be subjective, multiple options were given to participants to correct scenarios. Participants could flag missed relevant flights as ``Some Relevancy'' or ``Highly Relevant'' and similarly mistakenly identified, irrelevant flights could be flagged as ``Some Relevancy'' or ``Not Relevant''. Including a range of options allowed us to capture any variability of opinion between the ATCOs and provided a method to consolidate the responses of the two ATCOs into a single ``truth'' value for which aircraft were relevant in a scenario. A schema was created whereby a single response for the major option (``Highly Relevant'' or ``Not Relevant'') or two responses for the minor option (``Some Relevancy'') meant a correction was necessary. The 50 ATCO-corrected scenarios were used as a set of labelled traffic samples to tune the parameters $d$, $\Delta T$, $d_{CPA}$ and $t_{CPA}$. The F-beta score with a value of $\beta = 1.5$ was used to evaluate performance in the tuning process to prioritise fewer false negative results at the risk of an increased number of false positives.

\begin{figure}[ht]
\centering
\includegraphics[width=0.99\columnwidth]{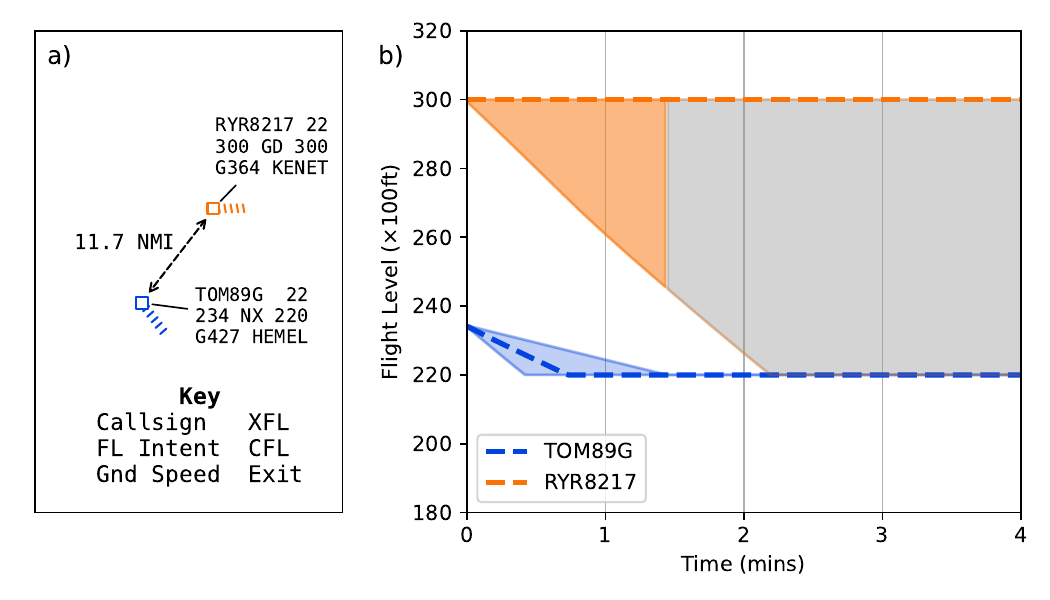}
\caption{Example showing the use of probabilistic TP in the Bluebird Digital Twin to determine feasible vertical profiles for aircraft. Panel a) shows a situation in which two aircraft are separated vertically but will cross tracks in the lateral plane. Panel b) shows a cone of uncertainty for the vertical profile of each aircraft over the following 4 minutes. The time period after the aircraft pass each other laterally (and are no longer crossing tracks) is illustrated in grey. As the span of current (FL), cleared (CFL) and exit (XFL) flight levels for each aircraft intersects, this pair would be classified as relevant by the original filter of Vijay Kumbhar~\textit{et al.}, however our data-driven modelling of climb and descent profiles determines this pair to be not relevant.}
\vspace{-1mm}
\label{fig:vert_TP}
\end{figure}

Following the first survey stage, ATCOs were invited to give qualitative feedback to the researchers to guide the adaptation of the relevant aircraft tool for LMS. As a result of these sessions, the Vijay Kumbhar~\textit{et al.} filter was extended in several respects: 

\begin{enumerate}
    \item \textbf{Handling in-trail aircraft}: A key traffic flow in LMS consists of aircraft beginning their descent into airports in the north of England. Many of these aircraft are placed under speed control by ATCOs. In such cases, aircraft may be within $d_{CPA}$ but not considered relevant as their relative velocity is negligible. Practically, this logic was implemented by evaluating whether an aircraft was within a $60^\circ$ lateral cone behind the subject aircraft and flying with the same or lower ground speed. Aircraft fulfilling these criteria were not considered relevant.
    \item \textbf{Diverging routes}: Aircraft following routes that diverged from one another were not treated as relevant provided there was at least 5 nautical miles (NM) of lateral separation at all times. 
    \item \textbf{Data-driven modelling of aircraft climb and descent}: The sole vertical consideration of the Vijay Kumbhar~\textit{et al.} filter for whether a pair of aircraft were relevant was if the span of the current, selected and sector exit flight levels of both aircraft intersected. This was found to be overly conservative in LMS, where most aircraft transiting this sector are either climbing to reach their cruising altitude or descending toward their destination. Many aircraft flying through LMS therefore occupy a wide span of flight levels and this single vertical component to the filter algorithm had little discriminative effect.
    
    To improve the filter in this regard, the probabilistic trajectory predictor in the Bluebird Digital Twin was used to estimate a cone of uncertainty for the vertical profile of each aircraft, based on the 2$\sigma$ credible interval. The filter used the level ranges from these cones (with a buffer of 10FL) to determine whether aircraft within the lateral threshold were relevant to one another, see Figure~\ref{fig:vert_TP}. This was found to give a more accurate prediction of relevancy than blocking all levels between the current and exit flight levels. The interested reader is referred to Hodgkin~\textit{et al.} \cite{hodgkin2025probabilisticsimulationaircraftdescent} for more details regarding the probabilistic trajectory predictor.

    \item \textbf{TP using operational data}: Finally, the relevant aircraft filter was adapted to move away from the state- and intent-based modes introduced by Vijay Kumbhar~\textit{et al.}. Instead, a single trajectory prediction was performed for each query aircraft. Leveraging the operational data available through the digital twin, these TP rollouts were performed based on the last issued ATCO instruction to each aircraft, better aligning the predicted trajectory with the operational situation within the sector. Therefore, the predicted trajectories for query aircraft either followed their planned route or proceeded along a fixed heading.
    
    Recognising that aircraft in LMS are often instructed to shortcut their route using either route directs or vectoring (see Fig. 1, Pepper~\textit{et al.} \cite{pepper4984556probabilistic}), two trajectories were generated for the selected subject aircraft in the scenario: one that followed the last issued clearance by an ATCO and a second trajectory that navigated directly to the final fix within LMS. These two trajectories, buffered by the distance $d_{CPA}$, swept out a portion of the sector into which the subject aircraft under consideration would likely be directed by ATCOs.  
\end{enumerate}

\begin{figure}[ht]
\centering
\includegraphics[width=0.93\columnwidth]{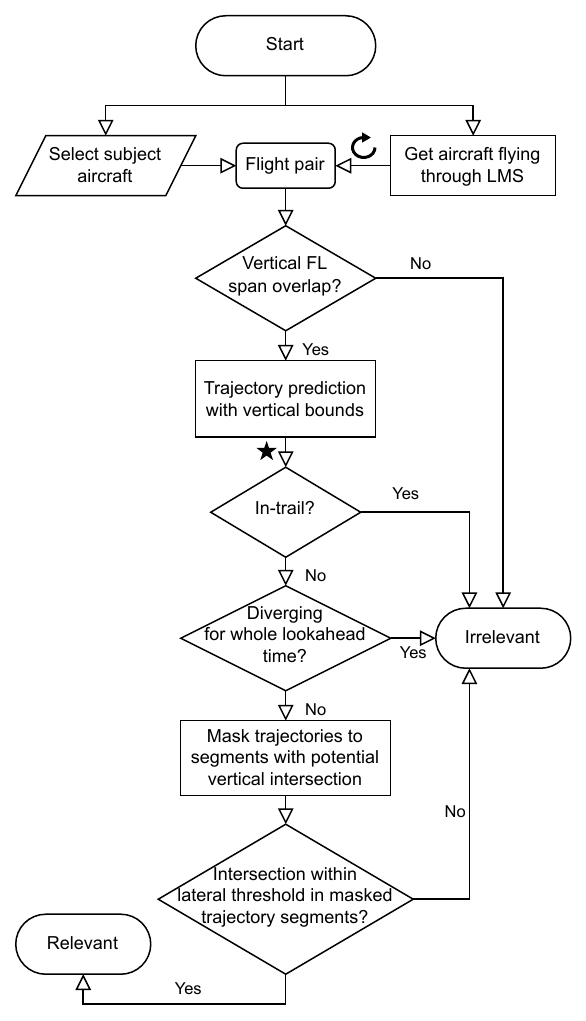}
\caption{Flowchart illustrating the outline of our updated relevant aircraft filtering method. The process is repeated for each pair of subject and other aircraft in LMS. At the star, the process is repeated for both subject aircraft trajectories, one following the last ATC instruction and the other navigating the aircraft directly to the exit of the sector.}
\label{fig:flowchart}
\end{figure}

Figure~\ref{fig:flowchart} displays our updated relevant aircraft filter in the form of a flowchart. After obtaining all flights within or due to enter LMS sector in the lookahead time (12 minutes), each aircraft is considered as a pair with the selected subject aircraft and the filtering process determines relevance. The star indicates a bifurcation point where the process is repeated for both of the trajectories for the subject aircraft (last ATCO instruction and route direct to exit), however a single path is shown for clarity of visualisation. A pair of aircraft is relevant if either pathway determines relevancy; this takes precedence over an irrelevant identification.

After the development phase was completed, the ATCOs were issued with a survey containing a final set of 50 new static traffic samples to assess the performance of our tuned and updated relevant aircraft filtering algorithm.

\subsection{Forecasting the number of relevant aircraft interactions}

The filter described in the previous sub-section provides a means of identifying pairs of relevant aircraft given a snapshot of traffic data. Through ATCO feedback it has been calibrated to operations within a specific sector of airspace. In this sub-section we describe in detail a method for predicting the number of relevant aircraft pairs within LMS using operational data, the proposed filter and a graph-based representation of LMS.

\subsubsection{Arrival time prediction using operational data}

Our method leverages the NATS National Airspace System Flight Data Processor (NAS FDP) to provide the complexity prediction model with livestreamed operational data. Messages from this system contain flight plan information for aircraft up to an hour before they reach the LMS sector and also include predictions of the time an aircraft will reach each waypoint on its planned route. Table~\ref{tab:livestream_message} shows an example excerpt of one of these NAS FDP messages and the data contained.

\begin{table}[ht]
\centering
\caption{\label{tab:livestream_message} An example excerpt of data from a NAS system message concerning a flight from Amsterdam to Bristol through the LMS sector.}
\begin{tabular}{cccc}
\textbf{Message Time} & \textbf{Callsign} & \textbf{Origin} & \textbf{Destination} \\
\noalign{\vspace{0.25ex}}
10:50:14 & KLM53Q & EHAM & EGGD \\
\noalign{\vspace{1ex}}
\multicolumn{4}{c}{\textbf{List of Waypoint Names}} \\
\noalign{\vspace{0.25ex}}
\multicolumn{4}{c}{{[}..., JACKO, MANGO, BRASO, LAM, POMAX, ...{]}} \\
\noalign{\vspace{1ex}}
\multicolumn{4}{c}{\textbf{Predicted Times at Waypoints}} \\
\noalign{\vspace{0.25ex}}
\multicolumn{4}{c}{{[}..., 11:33:00, 11:35:30, 11:35:54, 11:38:00, 11:49:42, ...{]}}
\end{tabular}
\end{table}

\begin{figure*}[ht]
\centering
\includegraphics[width=0.9\linewidth]{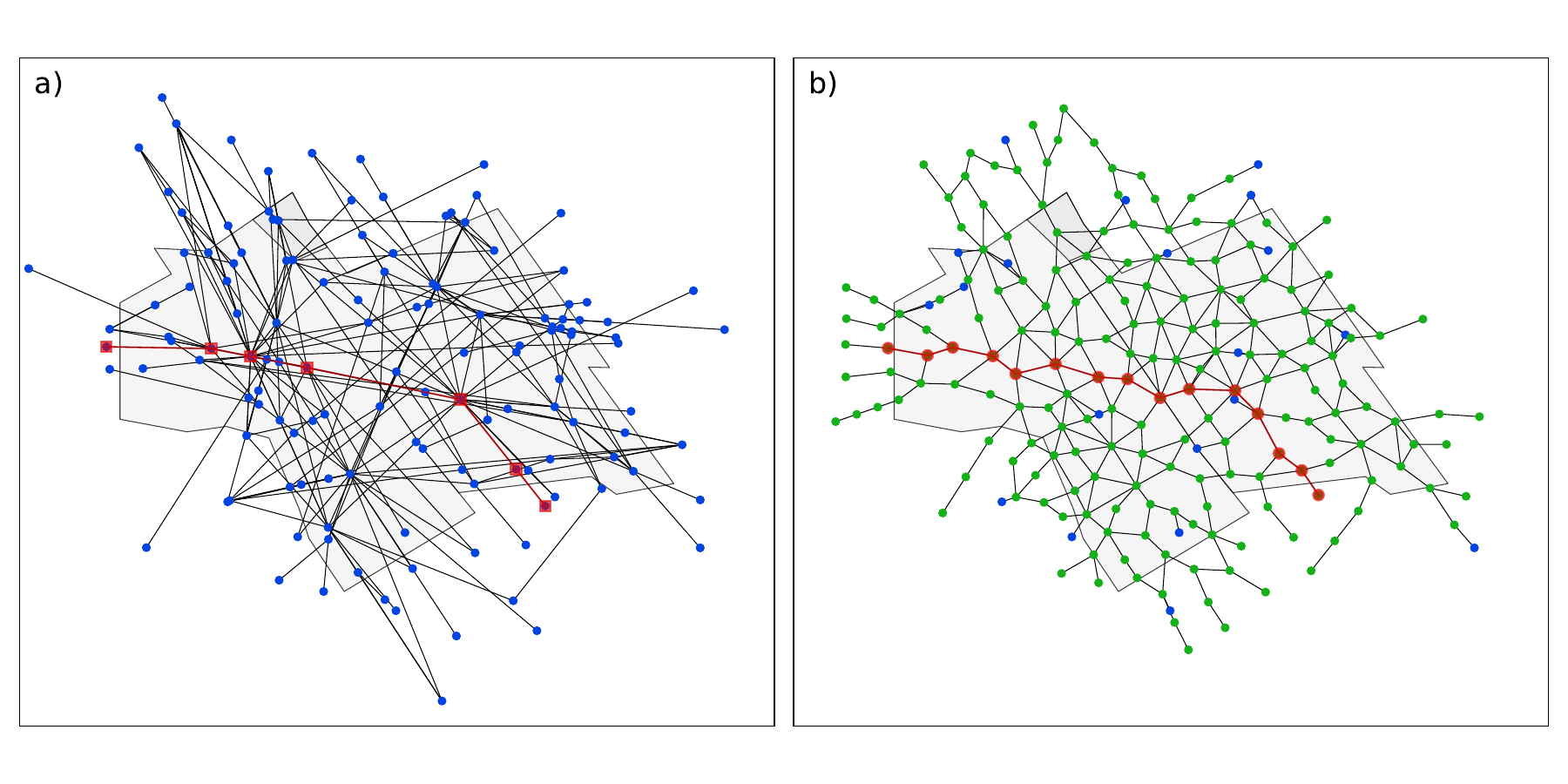}
\caption{a) The original graph created from the routes between waypoints (blue nodes) aircraft can take through LMS, the outline of which is shown in the background. The route structure appears a little chaotic as there are many different combinations of waypoints used for similar paths through the airspace. An example individual route for an aircraft flying through the sector is shown with red squares. b) The resampled route structure graph with newly introduced waypoints shown in green. The spatial resolution is more consistent with all edges between 5-10~NM in length. The corresponding path through the resampled graph for the example route shown in a) is highlighted in red.}
\label{fig:resampled_graph}
\end{figure*}

Access to the NAS FDP system has the advantage of providing the proposed complexity prediction tool access to updates regarding aircraft arrival time in LMS and allows the model to account for delays to an aircraft's arrival within the sector in real-time. However, a complication to using data from this system is that the separation between waypoints is significantly larger than the threshold lateral distance, $d_{CPA}$, used to determine relevancy. Waypoints are irregularly spaced and some flights following similar trajectories may have flight plans that include or omit different sets of waypoints. To rectify these issues, we convert the routes used by aircraft flying through LMS into a graph structure with standardised spatial fidelity.

\subsubsection{A graph-based route representation}

To create a graph of routes flown through LMS, we collected data for all aircraft that flew through the sector between January and December 2024. The filed flight plan for each flight was extracted and added into a single graph where nodes represented waypoints and edges were created between consecutive waypoints on a route. Routes were truncated at a waypoint before or after those that were laterally contained by the LMS boundary. This resulted in an initial graph of historic routes with 556 nodes and 1356 edges.

The graph was then refined through linear subdivision of edges (with a maximum final edge length of 5~NM) and an iterative process of agglomerative hierarchical clustering~\cite{Mullner2011} to obtain a resampled structure with route legs that were between 5-10~NM in length. The resampled graph contained 365 nodes and 980 edges. A look-up table was created to translate routes using the original waypoints to the resampled structure. The benefits of this approach were twofold: firstly it allowed us to collate flights with similar routes to a consistent path through the resampled graph; secondly it allowed predictions of aircraft location with finer and more regular spatial fidelity. Figure~\ref{fig:resampled_graph} shows the original graph construction of routes flown through LMS and our resampled graph structure. Sparse versions of the two graphs containing a subset of the nodes and edges are shown for clarity of visualisation. An example route flown through LMS is highlighted in red and the corresponding version is also shown on the resampled graph.

\subsubsection{Estimating aircraft location from waypoint arrival times}
We used the following statistical approach to estimate the probabilities of an aircraft currently occupying each route segment. $T_i$ is a random variable representing the arrival time at a waypoint $i$, where $i \in \{0,1,...,n-1\}$. The arrival times are assumed to follow normal distributions $T_i \sim \mathcal{N}(\mu_i, \sigma^2)$ and the probability that the aircraft has passed a waypoint $i$ at time $t$ is given by the Cumulative Distribution Function, $\Phi_i(t) = P(T_i \leq t)$. In this study a constant value of five minutes was determined suitable for $\sigma$ for all waypoints following initial empirical analysis of aircraft predicted and actual arrival times. In this setup, the probability of occupying each segment, $k$, at a specified query time, $t$, is given by:

\begin{equation}
P(L_k, t) = 
\begin{cases} 
1 - \Phi_0(t) & k = 0 \\
\Phi_{k-1}(t) - \Phi_{k}(t) & 1 \leq k < n \\
\Phi_{n-1}(t) & k = n
\end{cases}.
\end{equation}

\begin{figure*}[ht]
\centering
\includegraphics[width=0.92\linewidth]{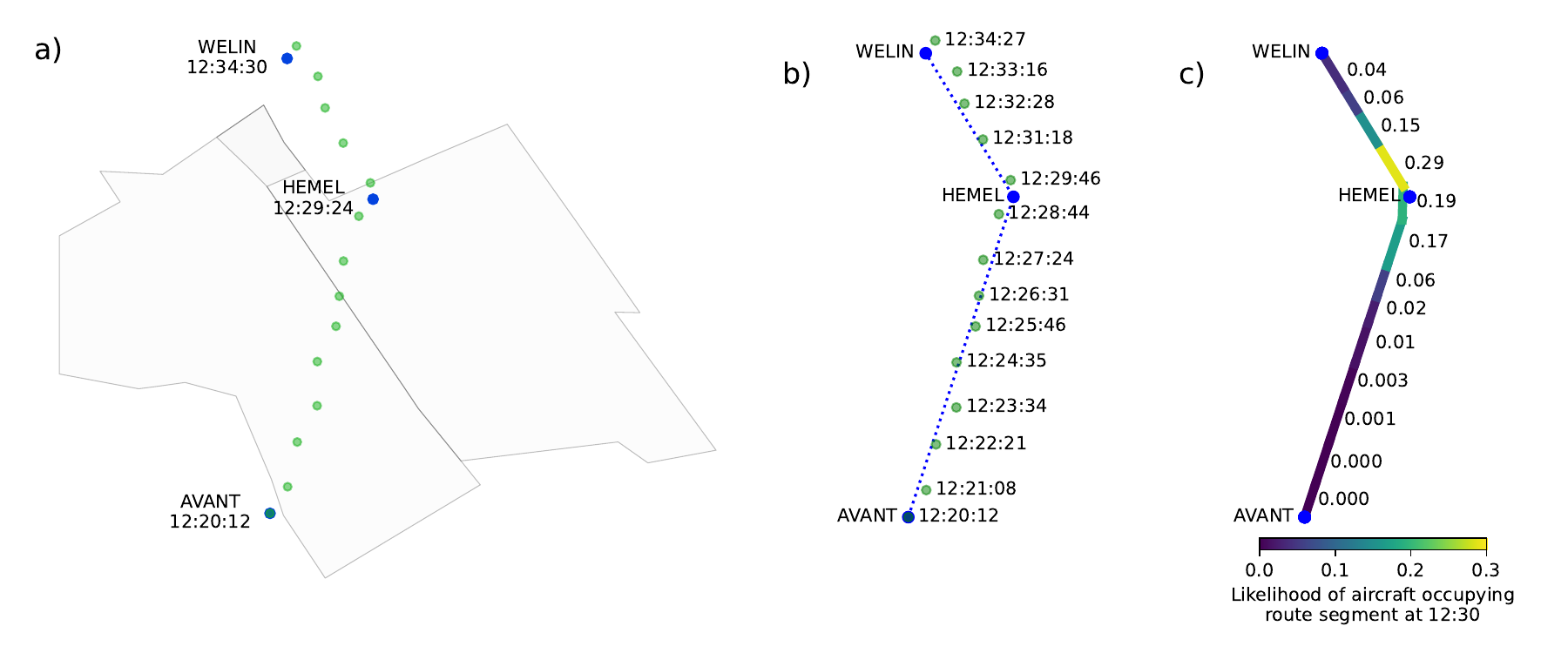}
\caption{An example of the procedure used to estimate the likelihood of an aircraft to occupy each leg along its route at a particular query time from the waypoint arrival time data. Panel a) shows a route planned through LMS with the original waypoints (shown in blue) annotated with the predicted arrival times. The corresponding waypoints on the resampled routes graph are shown in green. b) Linear interpolation is used to estimate the time the aircraft will arrive at the resampled waypoints. c) The likelihood of the aircraft occupying each segment between resampled waypoints at a query time of 12:30 is computed from the interpolated arrival times.}
\label{fig:interp}
\end{figure*}

A worked example of the procedure to estimate the probability of an aircraft occupying particular route legs in the resampled graph is demonstrated in Figure~\ref{fig:interp}.

\subsubsection{Complexity prediction}
In order to forecast the number of relevant pairs at a future query time, we gathered historic prior samples of relevant aircraft interactions and counts of which pairs of route legs were involved. 

Static LMS traffic scenarios were sampled every two minutes from January to December 2024. Our updated filter was used to evaluate each scenario, cycling through each aircraft in the sector as the subject.
We recorded each instance where two aircraft occupied a pair of route legs in our resampled graph structure of LMS and whether they had been identified as a relevant interacting pair. An individual historic prior was collected for each month to account for seasonal changes in traffic volume and patterns. The empirical expectation value for the number of relevant interactions given two aircraft occupying a pair of route legs $(\mathcal{O}_i, \mathcal{O}_j)$ in our resampled graph structure was then computed as

\begin{equation}
E[X \mid \mathcal{O}_i, \mathcal{O}_j] =  \frac{N(\text{relevant} \cap \mathcal{O}_i, \mathcal{O}_j)}{N(\mathcal{O}_i, \mathcal{O}_j)}.
\end{equation}

Despite the large number of flights considered, several pairs of legs had no historical observations, in which case the expectation term was set to zero to avert division errors.

At prediction time, to estimate the total expected number of relevant aircraft pairs we computed the following summation

\begin{equation}
E_{\text{Total relevant pairs}} = \sum_{i \in \mathcal{L}} \sum_{j \in \mathcal{L}} P(\mathcal{O}_i) P(\mathcal{O}_j) \cdot E[X \mid \mathcal{O}_i, \mathcal{O}_j],\label{eq}
\end{equation}

\noindent where $P(\mathcal{O}_i)$ is the likelihood of an aircraft occupying route leg $i$. The performance of the presented forecasting technique was evaluated over a day of data from May 2025, forecasting the number of relevant pairs at 30- and 45-minute lookahead times every minute between 06:00 and 16:00. 
This was compared against a direct approach to estimate the number of aircraft in the sector, or traffic volume, using the waypoint arrival time predictions. The Spearman's rank correlation coefficient (Spearman's $\rho$) was used to assess the monotonic relationship between forecast and observed values of relevant aircraft pairs. A moving block bootstrap approach was utilised to account for serial autocorrelation that was probable as traffic scenarios evolved continuously over the time series~\cite{Kunsch1989}. A total of 10,000 bootstrap samples were generated with a block size of 15 minutes. The difference in Spearman's $\rho$ was computed for each bootstrap sample, allowing us to construct a $95\%$ Confidence Interval (CI) and two-sided p-value.

\section{Results}
\subsection{Relevant aircraft filter adaptation}
\begin{figure}[h!]
\centering
\includegraphics[width=0.98\linewidth]{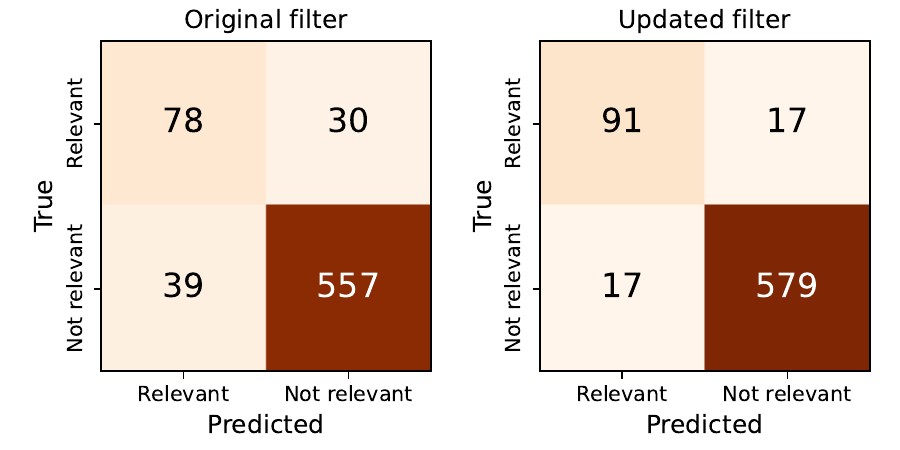}
\caption{Confusion matrices showing the performance of the filter of Vijay Kumbhar~\textit{et al.} with their original parameters and our updated approach assessed on 50 ATCO-labelled LMS traffic samples.}
\label{fig:res_confusion_matrix}
\end{figure}

To summarise the performance of both the original relevant aircraft filter of Vijay Kumbhar~\textit{et al.} and our updated approach on the final set of 50 ATCO-labelled LMS traffic samples, confusion matrices are displayed in Figure~\ref{fig:res_confusion_matrix}. The true label shows the ATCO-corrected class for each flight pair in the samples (relevant and not relevant) and the predicted label shows whether each filtering approach correctly identified flights in the scenario with respect to the subject aircraft. Counts aligning on the leading diagonal show agreement between the ATCOs and the filtering method. Precision, recall, F1-score and accuracy classification metrics are shown in Table~\ref{tab:comparison_results}. The updated filter outperforms the filter of Vijay Kumbhar~\textit{et al.} with their original parameters for every metric, highlighting the importance of adapting the approach to make it suitable for use in LMS.

\begin{figure*}[ht!]
\centering
\includegraphics[width=0.95\linewidth]{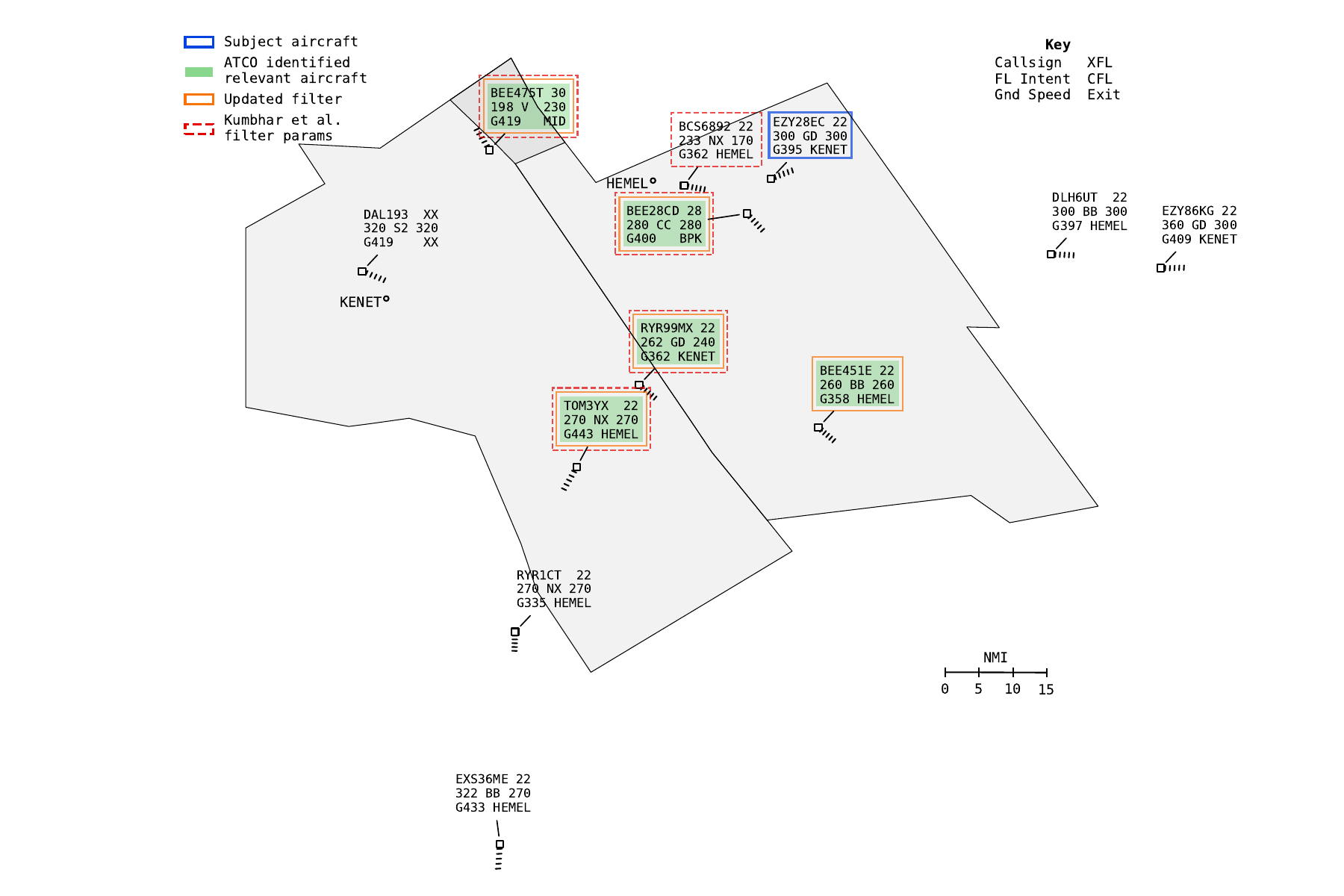}
\caption{Example traffic sample from the set of 50 ATCO-labelled scenarios of LMS. The subject aircraft is highlighted in blue and the ATCO-selected relevant aircraft are highlighted in green. Aircraft selected as relevant by the Vijay Kumbhar~\textit{et al.} filter are indicated with dashed red boxes. Flights selected as relevant by our updated approach are shown with an orange box. The updated filtering approach correctly identifies all relevant aircraft, whereas the original filter fails to flag aircraft BEE451E and incorrectly identifies the flight BCS6892 as relevant.}
\label{fig:filter_example}
\end{figure*}

\begin{table}[ht]
\centering
\caption{\label{tab:comparison_results}
Classification metrics comparing the performance of the Vijay Kumbhar~\textit{et~al.} filter with original parameters and our proposed updated relevant aircraft filter.}
\begin{tabularx}{0.97\columnwidth}{>{\centering\arraybackslash}X c c c c}
Filter & Precision & Recall & F1-score & Accuracy \\ \hline
\noalign{\vspace{1ex}}
Vijay Kumbhar~\textit{et~al.} filter and params & 0.67 & 0.72 & 0.69 & 0.90 \\
Our updated method & 0.84 & 0.84 & 0.84 & 0.95
\end{tabularx}
\end{table}

An example traffic sample from the evaluation set annotated by ATCOs is shown in Figure~\ref{fig:filter_example}. In this scenario the subject aircraft, highlighted in blue and with callsign EZY28EC, is flying westwards across the sector and shall descend from FL300 to FL220 as it approaches Bristol. The aircraft identified as relevant to the subject aircraft by the filter of Vijay Kumbhar~\textit{et al.} and our updated method are indicated with red and orange boxes respectively. The ATCO-identified relevant aircraft are shown with a green highlight. In this example, our updated filter approach is in full agreement with the ATCOs' selection of aircraft relevant to the subject. The original filter fails to identify an aircraft travelling north-west, with the callsign BEE451E, which the ATCOs determined to be relevant. Additionally, the original filter incorrectly identified aircraft BCS6892 as relevant. The route of BCS6892 crosses that of the subject aircraft but the aircraft have significant vertical separation (6700 feet) and both the lateral and vertical separation will grow between the aircraft as they proceed along their routes. Our updated filter correctly recognises BCS6892 as irrelevant to the subject due to the consideration of each aircraft's vertical profile using bounded trajectory prediction (outlined in section~\ref{meth:adaptation} and Figure~\ref{fig:vert_TP}). The waypoints named KENET and HEMEL are indicated in the figure to show the locations where the subject aircraft, EZY28EC, and BCS6892 are due to leave LMS.

\subsection{Complexity forecasting}

\begin{figure*}[ht]
\centering
\includegraphics[width=0.99\linewidth]{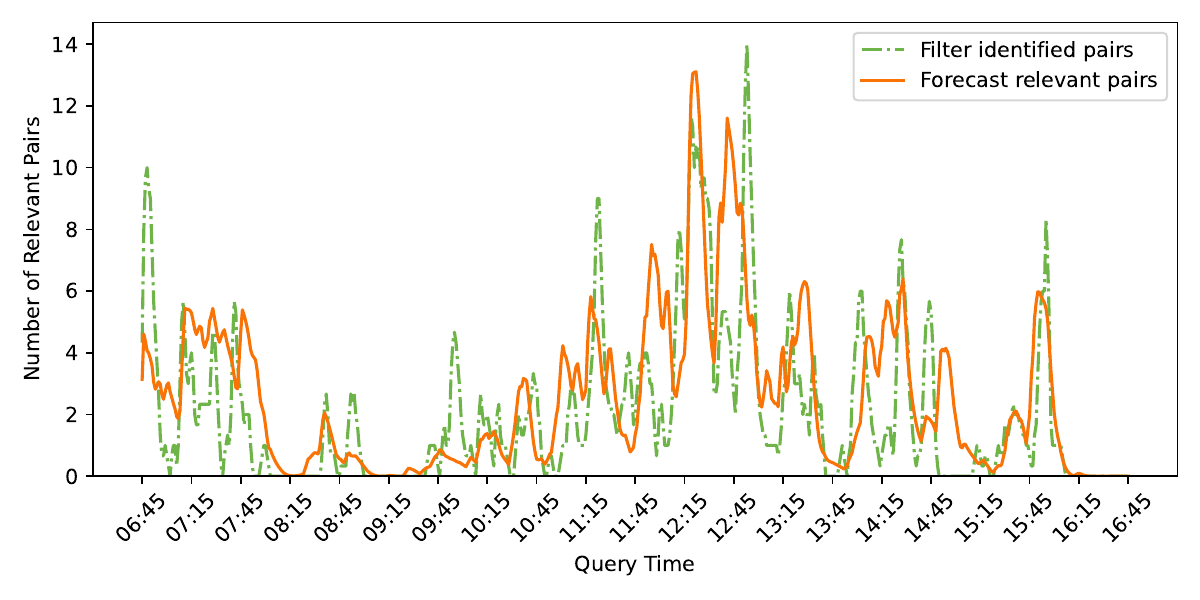}

\caption{Plot showing the number of relevant pairs forecast 45 minutes in advance by our proposed method (orange) for a day in May 2025. A forecast prediction was made every minute between 06:00 and 16:00. The actual number of pairs identified by our updated filter approach at the query times are shown in green. Overall the forecast predictions show good alignment with the number of pairs identified by our filter when the query time is reached.}

\label{fig:forecast_example}
\end{figure*}

The proposed complexity forecasting method to estimate the number of relevant pairs in future using NAS FDP data was evaluated on livestream data from a single day in May 2025. Forecast predictions were produced at 1-minute intervals between 06:00 and 16:00 at two selected lookahead times of 30 and 45 minutes. The actual number of relevant pairs, as identified by our updated filter method, was captured retrospectively from the dataset containing the radar, flight plan and clearances when the query time was reached. The constant value of $\sigma = 5$ minutes used to model aircraft occupancy of route legs was empirically found to be robust in this study, but future work ought to investigate the use of distance-dependent $\sigma$ values, as uncertainties are likely to accrue further along the future trajectory of a flight. Similarly, alternative modelling approaches for the arrival times, beyond the current normal distribution assumption, should be investigated in future.

Figure~\ref{fig:forecast_example} shows a plot of the number of relevant pairs forecast with a 45-minute lookahead (orange) and the number identified by our filter post hoc (green) across the span of the evaluation day. The peaks and troughs are mostly in good alignment, with the busiest period between 12:00-13:00 seeing a corresponding increase in the forecast number of relevant pair interactions. Numerically, the Spearman's $\rho$ between the forecast and filter-identified number of relevant pairs was 0.68 for the 30- and 45-minute lookahead times, indicating a moderate to strong positive relationship. We additionally used the predicted waypoint arrival times to directly forecast the number of aircraft in the sector at each lookahead time. The Spearman's $\rho$ was calculated between the forecast number of aircraft and the filter-identified number of relevant pairs as 0.54 and 0.55 for the 30- and 45-minute lookahead times respectively.
Our proposed forecasting method demonstrated stronger correlation with the observed number of relevant aircraft pairs than the direct prediction of traffic volume. For the 45-minute lookahead time, the moving block bootstrap analysis confirmed a statistically significant difference between the two approaches, with a mean bootstrapped difference of 0.134 and $95\%$ CI [0.064, 0.211]. As the $95\%$ CI is strictly positive, our approach significantly outperformed the direct prediction of traffic volume to forecast a proxy measure of ATCO workload (p=0.0004). Comparable results were observed for the 30-minute lookahead time (p=0.004). However, evaluation of the method was limited to a single day and so future work will need to assess the performance of the approach across different seasons and patterns of air traffic.

\section{Conclusion}

In this study, we presented a probabilistic framework to forecast airspace complexity based on the concept of relevant aircraft pairs instead of sector occupancy counts. We adapted and refined an existing relevant aircraft filtering algorithm through iterative engagement with licensed ATCOs to capture the specific geometric and operational characteristics of LMS. Our updated filter was shown to identify aircraft relevant to a selected subject with performance that was in strong agreement with the ATCOs (F1-score = 0.84). 
The tuned filter was integrated with a graph-based representation of the route network of LMS and a live operational data feed to facilitate predictions of the upcoming number of relevant pairs up to 45 minutes in advance. The proposed method had a significantly stronger correlation with the actual number of relevant pairs (Spearman's $\rho = 0.68$) compared to a standard forecast of the upcoming traffic volume in the sector ($\rho = 0.55$). This improvement shows the potential of the method as a tool for group supervisors, supporting decision making for sector configuration changes and ATCO scheduling based on the predicted number of relevant interactions instead of traffic count alone. Further evaluation on real-time live operational data, alongside careful consultation with ATCO group supervisors, will be required to demonstrate the tools operational impact.

The complexity forecasting tool is fast, producing predictions up to 45 minutes in advance in less than a second on a laptop. This is an encouraging sign that the method would be feasible for use in real-time, as predictions for multiple sectors could be made well within a single radar sweep ($\sim6$ seconds). 

The study presented is currently limited to a single sector of LACC. Further calibration and evaluation will be required to extend the relevant aircraft filter effectively to other sectors. Application of the complexity forecasting tool more widely to further sectors of airspace will require extension of the graph-based representation of the route network. Currently, the graph representation is 2D, relying on implicitly embedded information in the historic prior regarding which pairs of route legs involve vertical interactions. Future work will aim to augment the vertical component of the graph representation to allow concurrent complexity forecasting for sectors above LMS in LACC (FL305+). Extension of the methodology to sectors beyond LMS will result in the evaluation on a wider range of scenarios to ensure the forecasting tool is robust in a diverse variety of traffic flows and operational procedures.

\section*{Acknowledgement}
The authors would like to thank Dewi Gould, Andrew Pace and Lee Benson for their time and effort put into this study.



\FloatBarrier
\balance
\printbibliography{}

\end{document}